\documentclass[11pt,a4paper,table]{article}
\usepackage[hyperref]{emnlp2018}
\usepackage{times}
\usepackage{latexsym}
\usepackage{url}
\usepackage{graphicx,epsfig}
\usepackage{algorithm}
\usepackage{algorithmic}
\usepackage{amssymb}
\usepackage{multirow}
\usepackage{verbatim}
\usepackage{scrextend}
\usepackage{float}
\usepackage{subfig}
\usepackage{array}
\newcommand*\rot{\rotatebox{90}}
\usepackage{tablefootnote}

\newcolumntype{C}[1]{>{\centering\let\newline\\\arraybackslash\hspace{0pt}}m{#1}}
\usepackage{tikz}
\usetikzlibrary{shapes,arrows}
\usepackage{amsmath,bm,times}
\usepackage{verbatim}
\usepackage{subfig}
\usepackage{float}
\usetikzlibrary{positioning}
\usepackage{tkz-graph}

\usetikzlibrary{decorations.pathreplacing,intersections}
\usepackage{wrapfig}
\usepackage{xcolor}
\usepackage{pbox}
\usepackage{footmisc}
\usepackage{bbding}

\aclfinalcopy 

\title{A Multi-task Ensemble Framework for Emotion, Sentiment and Intensity Prediction}

\author{Md Shad Akhtar$^{\dagger}$\Envelope, Deepanway Ghosal$^\dagger$, Asif Ekbal$^\dagger$, Pushpak Bhattacharyya$^\dagger$, Sadao Kurohashi$^+$ \\
  $^\dagger$\textit{Department of Computer Science \& Engineering, Indian Institute of Technology Patna, India.} \\
  $^+$\textit{Department of Intelligence Science and Technology, Kyoto University, Japan.} \\
  {\tt \Envelope shad.pcs15@iitp.ac.in} 
  }

\date{}

\begin{document}
\maketitle
\begin{abstract}
In this paper, through multi-task ensemble framework we address three problems of emotion and sentiment analysis i.e. ``emotion \textit{classification} \& \textit{intensity}", ``\textit{valence}, \textit{arousal} \& \textit{dominance} for emotion" and ``\textit{valence} \& \textit{arousal} for sentiment". The underlying problems cover two granularities (i.e. \textit{coarse-grained} and \textit{fine-grained}) and a diverse range of domains (i.e. \textit{tweets}, \textit{Facebook posts}, \textit{news headlines}, \textit{blogs}, \textit{letters} etc.). The ensemble model aims to leverage the learned representations of three deep learning models (i.e. CNN, LSTM and GRU) and a hand-crafted feature representation for the predictions. Experimental results on the benchmark datasets show the efficacy of our proposed multi-task ensemble frameworks. We obtain the performance improvement of 2-3 points on an average over single-task systems for most of the problems and domains. 
\end{abstract}

\section{Introduction}\label{sec:intro}
Emotion \cite{Picard:1997:AC:265013} and sentiment \cite{Panget.al2005} are closely related and are often been used interchangeably. However, according to \newcite{munezero2014they}, emotions and sentiments differ on the scale of duration on which they are experienced. 
These have applications in a diverse set of real-world problems such as stock market predictions, disaster management systems, health management systems, feedback systems for an organization or individual user \textit{w.r.t.} a product or service to take an informed decision \cite{hawn2009take,bollen2011twitter,neubig2011safety}. Any organization does not wish to lose their valuable customers. They can keep track of varying emotions and sentiments of their customers over a period of time. If the unpleasant emotions or sentiments of the customer are increasing day-by-day, the organization can act in a timely manner to address his/her concerns. On the other hand, if the emotions and sentiments are pleasant the organization can ride on the positive feedbacks of their customers to analyze and forecast their economic situation with more confidence.

The classification of emotions and sentiments into coarse-grained classes does not always reflect exact state of mood or opinion of a user, hence, do not serve the purpose completely. Recently, the attention has been shifted towards fine-grained analysis on the continuous scale \cite{semEval-2007:affective-texts,fbpost:WASSA2016,emobank:2017,mohammad:WASSA2017,sentiment:intensity:mohammad:semeval:2016}. 
Arousal or intensity defines the degree of emotion and sentiment felt by the user and often differs on a case-to-case basis. Within a single class (e.g. \textit{Sadness}) some emotions are gentle (e.g `\textit{I lost my favorite pen today.}') while others can be severe (e.g. `\textit{my uncle died from cancer today...RIP}'). Similarly, some sentiments are gentler than others within the same polarity, e.g. `\textit{happy to see you again}' v/s `\textit{can't wait to see you again}'. 

In our current research, we aim to solve these inter-related problems i.e. ``\textit{emotion classification \& intensity}" for coarse-grained emotion classification, ``\textit{valence, arousal \& dominance}" for fine-grained emotion analysis and ``\textit{valence \& arousal}" for fine-grained sentiment analysis\footnote{\textit{Fine-grained} refers to the prediction on a continuous scale, whereas \textit{coarse-grained} refers to the prediction on a discrete level \cite{semEval-2007:affective-texts,emobank:2017}.}. We propose an efficient multi-task ensemble framework that tackles all these problems concurrently.

Multi-task learning framework targets to achieve generalization by leveraging the inter-relatedness of multiple problems/tasks \cite{Caruana1997}. The intuition behind multi-task learning is that if two or more tasks are correlated then the joint-model can learn effectively from the shared representations. In comparison to the single-task framework, where different tasks are solved separately, a multi-task framework offers three main advantages i.e. a) achieves better generalization; b) improves the performance of each task through shared representation; and c) requires only one unified model in contrast to separate models for each task in single-task setting, resulting in reduced complexity. 

Our proposed multi-task framework is greatly inspired from this, and it jointly performs multiple tasks. Our framework is based on an ensemble technique. At first, we learn hidden representations through three deep learning models, i.e. Convolutional Neural Network (CNN), Long Short Term Memory (LSTM) and Gated Recurrent Unit(GRU). We subsequently feed the learned representations of three deep learning systems along with a hand-crafted feature vector to a Multi-Layer Perceptron (MLP) network to construct an ensemble. The objective is to leverage four different representations and capture the relevant features among them for the predictions. The proposed network aims to predict multiple outputs from the input representations in one-shot.

We evaluate the proposed approach for three problems i.e. \textit{coarse-grained emotion analysis}, \textit{fine-grained emotion analysis} and \textit{fine-grained sentiment analysis}. For \textit{coarse-grained emotion analysis}, we aim to predict emotion class and its intensity value as the two tasks. The first task (i.e. \textit{emotion classification}) classifies the incoming tweet into one of the predefined classes (e.g. \textit{joy}, \textit{anger}, \textit{sadness}, \textit{fear} etc.) and subsequently the second task (i.e. \textit{emotion intensity prediction}) predicts the associated degree of emotion expressed by the writer in a continuous range of 0 to 1. 
In \textit{fine-grained emotion analysis}, we aim to predict the \textit{valence}, \textit{arousal} and \textit{dominance} scores in parallel, whereas, in the third problem, i.e. \textit{fine-grained sentiment analysis}, our goal is to predict \textit{valence} and \textit{arousal} scores in a multi-task framework. The range of each task of the second and third problems is on the continuous scale of \textit{1} to \textit{5} and \textit{1} to \textit{9}, respectively. In total, we apply the proposed multi-task approach for three configurations: a) multi-tasking for classification (\textit{emotion classification}) and regression (\textit{emotion intensity prediction}) together; b) multi-tasking for two regression tasks together (sentiment \textit{valence} \& \textit{arousal} prediction); and c) multi-tasking for three regression tasks together (emotion \textit{valence}, \textit{arousal} \& \textit{dominance}).  

The main contributions of our proposed work are summarized below: \textbf{a)} we effectively combine deep learning representations with manual features via an ensemble framework; and \textbf{b)} we develop a multi-task learning framework which attains overall better performance for different tasks related to emotion, sentiment and intensity. 

\section{Related Works}
\label{sec:lit}
Literature suggests that multi-task learning has been successfully applied in a multitude of machine learning (including natural language processing) problems \cite{multitask:nlp:2008:icml,multitask:nlp:2016:acl,multitask-sentiment:2017:sigir,multitask-emotion-ieee-affective}. Authors in \cite{multitask-sentiment:2017:sigir} employed recurrent neural network for their multi-task framework where they treated \textit{3-way} classification and \textit{5-way} classification as two separate tasks for sentiment analysis. One of the earlier works on emotion detection looks at emotion bearing words in the text for classification \newcite{ekman1999}. In another work, \newcite{dungEmotion} studied human mental states \textit{w.r.t.} an emotion for training a Hidden Markov Model (HMM). These systems concentrated on emotion classification, whereas, the works reported in \cite{mohammad:WASSA2017,jain-EtAl:2017:WASSA2017,koper-kim-klinger:2017:WASSA2017} focus only on intensity prediction. \newcite{jain-EtAl:2017:WASSA2017} used an ensemble of five different neural network models for predicting the emotion intensity. They also explored the idea of multi-task learning in one of the models, where they treated four different emotions as the four tasks. The final predictions were generated by a weighted average of the base models. \newcite{koper-kim-klinger:2017:WASSA2017} employed a random forest regression model on the concatenated lexicon features and CNN-LSTM features. Authors in \cite{akhtar-EmoInt17} employed LSTM and SVR in cascade for predicting the emotion intensity. Recently, \newcite{multitask-emotion-ieee-affective} have proposed VA (\textit{Valence-Activation}) model for emotion recognition in 2D continuous space. Following the trends of emotion intensity prediction, researchers have also focused on predicting the intensity score for sentiment \cite{sentiment:intensity:balahur:2011,sentiment:intensity:mohammad:semeval:2016,sentiment:intensity:sharma:2017:emnlp,akhtar:MLP:EMNLP2017}. 

Traditional techniques e.g. Boosting \cite{Freund1996}, Bagging \cite{Breiman1996}, Voting (Weighted, Majority) \cite{Kittler:1998:CC:279005.279007} etc. are some of the common choices for constructing ensemble \cite{Ekbal:2011:WVC:1967293.1967296,Ens-Bagging-Boosting,Ens-majority}. Recently, \newcite{Akhtar:KBS2017116} proposed an ensemble technique based on Particle Swarm Optimization (PSO) to solve the problem of aspect based sentiment analysis. 

Our proposed approach differs with these existing systems in terms of the following aspects: a) MLP based ensemble addresses both \textit{classification} and \textit{regression} problems; b) multi-task framework handles diverse set of tasks (i.e. \textit{classification \& regression problems, 2 regression problems} and \textit{3 regression problems}); and c) our proposed approach covers two granularities (i.e. coarse-grained \& fine-grained) and a diverse set of domains (i.e. \textit{tweets, fb posts, news headlines, blogs, letters} etc.).

\section{Proposed Methodology}
\label{sec:method}
Ensemble is an efficient technique in combining the outputs of various candidate systems. The basic idea is to leverage the goodness of several systems to improve the overall performance. Motivated by this, we propose a multi-task ensemble learning framework built on top of learned representations of three deep learning models and a hand-crafted feature vector. We separately train all three deep learning models, i.e. a CNN, a LSTM and a GRU network in a multi-task framework (Figure \ref{fig:approach:ind}). Once the network is trained, we extract an intermediate layer activation from these CNN, LSTM and GRU models. These three \textit{task-aware} deep representations are concatenated with a feature vector before feeding into the multi-task ensemble model. The multi-task ensemble model is a MLP network which comprises of four hidden layers. The first two hidden layers are shared for all the tasks and the final two hidden layers are specific for each individual task. The idea is to exploit the goodness of different feature representations and to learn a combined representation for solving multiple tasks. Consequently, we show that the ensemble model performs better than each of the individual models. A high-level outline of the proposed approach is depicted in Figure \ref{fig:approach}. Figure \ref{fig:approach:ind} shows the multi-task framework for the individual CNN, LSTM and GRU models. After training, the respective \textit{task-aware} intermediate representations (\textit{color coded green} in Figure \ref{fig:approach:ind}) and the hand-crafted feature vector are used as input for the ensemble in Figure \ref{fig:approach:ens}. 
\tikzstyle{in-out}=[draw, fill=red!20, text width=15em, outer sep=2, text centered, minimum height=3em, rounded corners]
\tikzstyle{in-out-large}=[draw, fill=green!30, text width=8em, outer sep=2, text centered, minimum height=5em, rounded corners]
\tikzstyle{process}=[draw, outer sep=2, fill=teal!20, text width=31.5em, text centered, minimum height=3em]
\tikzstyle{process-2}=[draw, outer sep=2, fill=green!30, text width=31.5em, text centered, minimum height=3em, rounded corners]
\tikzstyle{process-small}=[draw, outer sep=2, fill=blue!20, text width=15em, text centered, minimum height=2.5em]
\tikzstyle{process-large}=[draw, outer sep=2, fill=blue!20, text width=31.5em, text centered, minimum height=4em]
\tikzstyle{process-huge}=[draw, outer sep=2, fill=blue!20, text width=38em, text centered, minimum height=6em]

\tikzstyle{in-out-new}=[draw, fill=red!20, text width=5em, outer sep=2, text centered, minimum height=3em, rounded corners]
\tikzstyle{process-2-new}=[draw, outer sep=2, fill=green!30, text width=12em, text centered, minimum height=3em, rounded corners]
\tikzstyle{process-small-new}=[draw, outer sep=2, fill=blue!20, text width=5em, text centered, minimum height=2.5em]
\tikzstyle{process-large-new}=[draw, outer sep=2, fill=blue!20, text width=12em, text centered, minimum height=4em]
\tikzstyle{process-new}=[draw, outer sep=2, fill=teal!20, text width=41.0em, text centered, minimum height=3em]

\tikzstyle{in-out-dummy}=[text width=3em, outer sep=0, text centered, minimum height=3em, rounded corners]

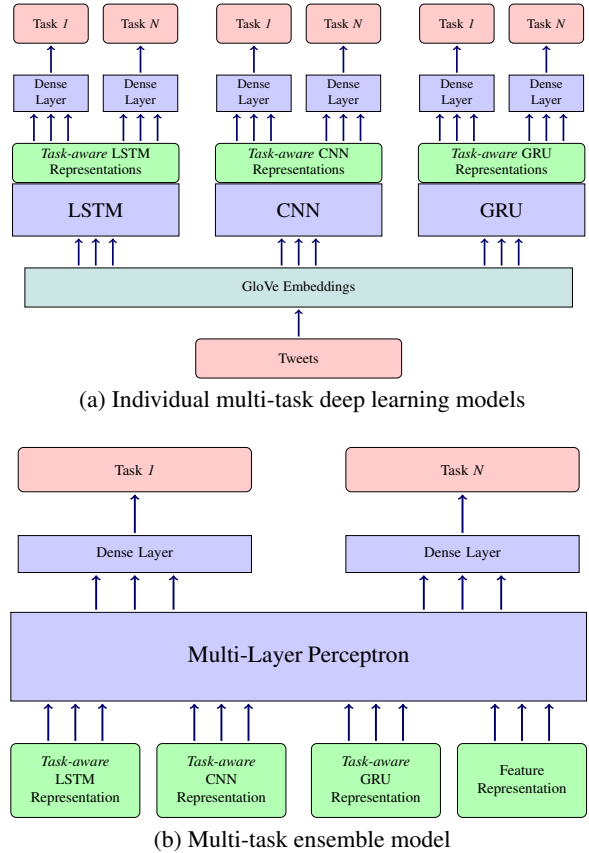
\begin{figure}[h!]
	\centering
{
	   	\subfloat[Individual multi-task deep learning models\label{fig:approach:ind}]
    	{%
        \centering
        \resizebox{0.48\textwidth}{!}{
			\begin{tikzpicture}
\node[in-out-new] (dl-class1) {Task \textit{1}};
\node[in-out-new, node distance=0.3cm, anchor=south] (dl-intensity1) [right=of dl-class1] {Task \textit{N}};

\node[in-out-new] (dl-class2) [right=of dl-intensity1] {Task \textit{1}};
\node[in-out-new, node distance=0.3cm, anchor=south] (dl-intensity2) [right=of dl-class2] {Task \textit{N}};

\node[in-out-new] (dl-class3) [right=of dl-intensity2] {Task \textit{1}};
\node[in-out-new, node distance=0.3cm, anchor=south] (dl-intensity3) [right=of dl-class3] {Task \textit{N}};

\node[process-small-new, node distance=0.8cm, anchor=south] (dl-class-dense1) [below=of dl-class1] {Dense Layer};
\node[process-small-new, node distance=0.8cm, anchor=south] (dl-intensity-dense1) [below=of dl-intensity1] {Dense Layer};

\node[process-small-new, node distance=0.8cm, anchor=south] (dl-class-dense2) [below=of dl-class2] {Dense Layer};
\node[process-small-new, node distance=0.8cm, anchor=south] (dl-intensity-dense2) [below=of dl-intensity2] {Dense Layer};

\node[process-small-new, node distance=0.8cm, anchor=south] (dl-class-dense3) [below=of dl-class3] {Dense Layer};
\node[process-small-new, node distance=0.8cm, anchor=south] (dl-intensity-dense3) [below=of dl-intensity3] {Dense Layer};

\path (dl-class-dense1.south west) -- (dl-intensity-dense1.south east) node[process-2-new] (dl-representation1) [midway, below=0.8cm] {\large \textit{Task-aware} LSTM Representations};
\path (dl-representation1.south west) -- (dl-representation1.south east) node[process-large-new] (dl-network1) [midway, below=-0.1cm] {\LARGE LSTM};

\path (dl-class-dense2.south east) -- (dl-intensity-dense2.south west) node[process-2-new] (dl-representation2) [midway, below=0.8cm] {\large \textit{Task-aware} CNN Representations};
\path (dl-representation2.south east) -- (dl-representation2.south west) node[process-large-new] (dl-network2) [midway, below=-0.1cm] {\LARGE CNN};

\path (dl-class-dense3.south east) -- (dl-intensity-dense3.south west) node[process-2-new] (dl-representation3) [midway, below=0.8cm] {\large \textit{Task-aware} GRU Representations};
\path (dl-representation3.south east) -- (dl-representation3.south west) node[process-large-new] (dl-network3) [midway, below=-0.1cm] {\LARGE GRU};

\path (dl-network1.south west) -- (dl-network3.south east) node[process-new] (dl-embed) [midway, below=0.8cm] {\large GloVe Embeddings};
\path (dl-embed.south east) -- (dl-embed.south west) node[in-out] (dl-tweet) [midway, below=0.8cm] {\large Tweets};

\path[line width=0.5mm, black!60!blue, ->] (dl-class-dense1) edge (dl-class1);
\path[line width=0.5mm, black!60!blue, ->] (dl-intensity-dense1) edge (dl-intensity1);

\path[line width=0.5mm, black!60!blue, ->] (dl-class-dense2) edge (dl-class2);
\path[line width=0.5mm, black!60!blue, ->] (dl-intensity-dense2) edge (dl-intensity2);

\path[line width=0.5mm, black!60!blue, ->] (dl-class-dense3) edge (dl-class3);
\path[line width=0.5mm, black!60!blue, ->] (dl-intensity-dense3) edge (dl-intensity3);

\path[line width=0.5mm, black!60!blue, ->] (dl-tweet) edge (dl-embed);

\path [draw, line width=0.5mm, black!60!blue, <-] (dl-class-dense1.south) -- (dl-representation1.north -| dl-class-dense1);
\path [draw, line width=0.5mm, black!60!blue, <-] ([xshift=0.5cm]dl-class-dense1.south) -- ([xshift=0.5cm]dl-representation1.north -| dl-class-dense1);
\path [draw, line width=0.5mm, black!60!blue, <-] ([xshift=-0.5cm]dl-class-dense1.south) -- ([xshift=-0.5cm]dl-representation1.north -| dl-class-dense1);

\path [draw, line width=0.5mm, black!60!blue, <-] (dl-intensity-dense1.south) -- (dl-representation1.north -| dl-intensity-dense1);
\path [draw, line width=0.5mm, black!60!blue, <-] ([xshift=0.5cm]dl-intensity-dense1.south) -- ([xshift=0.5cm]dl-representation1.north -| dl-intensity-dense1);
\path [draw, line width=0.5mm, black!60!blue, <-] ([xshift=-0.5cm]dl-intensity-dense1.south) -- ([xshift=-0.5cm]dl-representation1.north -| dl-intensity-dense1);


\path [draw, line width=0.5mm, black!60!blue, <-] (dl-intensity-dense2.south) -- (dl-representation2.north -| dl-intensity-dense2);
\path [draw, line width=0.5mm, black!60!blue, <-] ([xshift=0.5cm]dl-intensity-dense2.south) -- ([xshift=0.5cm]dl-representation2.north -| dl-intensity-dense2);
\path [draw, line width=0.5mm, black!60!blue, <-] ([xshift=-0.5cm]dl-intensity-dense2.south) -- ([xshift=-0.5cm]dl-representation2.north -| dl-intensity-dense2);

\path [draw, line width=0.5mm, black!60!blue, <-] (dl-class-dense2.south) -- (dl-representation2.north -| dl-class-dense2);
\path [draw, line width=0.5mm, black!60!blue, <-] ([xshift=0.5cm]dl-class-dense2.south) -- ([xshift=0.5cm]dl-representation2.north -| dl-class-dense2);
\path [draw, line width=0.5mm, black!60!blue, <-] ([xshift=-0.5cm]dl-class-dense2.south) -- ([xshift=-0.5cm]dl-representation2.north -| dl-class-dense2);


\path [draw, line width=0.5mm, black!60!blue, <-] (dl-class-dense3.south) -- (dl-representation3.north -| dl-class-dense3);
\path [draw, line width=0.5mm, black!60!blue, <-] ([xshift=0.5cm]dl-class-dense3.south) -- ([xshift=0.5cm]dl-representation3.north -| dl-class-dense3);
\path [draw, line width=0.5mm, black!60!blue, <-] ([xshift=-0.5cm]dl-class-dense3.south) -- ([xshift=-0.5cm]dl-representation3.north -| dl-class-dense3);

\path [draw, line width=0.5mm, black!60!blue, <-] (dl-intensity-dense3.south) -- (dl-representation3.north -| dl-intensity-dense3);
\path [draw, line width=0.5mm, black!60!blue, <-] ([xshift=0.5cm]dl-intensity-dense3.south) -- ([xshift=0.5cm]dl-representation3.north -| dl-intensity-dense3);
\path [draw, line width=0.5mm, black!60!blue, <-] ([xshift=-0.5cm]dl-intensity-dense3.south) -- ([xshift=-0.5cm]dl-representation3.north -| dl-intensity-dense3);

\path [draw, line width=0.5mm, black!60!blue, <-] (dl-network3.south) -- (dl-embed.north -| dl-network3);
\path [draw, line width=0.5mm, black!60!blue, <-] ([xshift=0.5cm]dl-network3.south) -- ([xshift=0.5cm]dl-embed.north -| dl-network3);
\path [draw, line width=0.5mm, black!60!blue, <-] ([xshift=-0.5cm]dl-network3.south) -- ([xshift=-0.5cm]dl-embed.north -| dl-network3);

\path [draw, line width=0.5mm, black!60!blue, <-] (dl-network2.south) -- (dl-embed.north -| dl-network2);
\path [draw, line width=0.5mm, black!60!blue, <-] ([xshift=0.5cm]dl-network2.south) -- ([xshift=0.5cm]dl-embed.north -| dl-network2);
\path [draw, line width=0.5mm, black!60!blue, <-] ([xshift=-0.5cm]dl-network2.south) -- ([xshift=-0.5cm]dl-embed.north -| dl-network2);

\path [draw, line width=0.5mm, black!60!blue, <-] (dl-network1.south) -- (dl-embed.north -| dl-network1);
\path [draw, line width=0.5mm, black!60!blue, <-] ([xshift=0.5cm]dl-network1.south) -- ([xshift=0.5cm]dl-embed.north -| dl-network1);
\path [draw, line width=0.5mm, black!60!blue, <-] ([xshift=-0.5cm]dl-network1.south) -- ([xshift=-0.5cm]dl-embed.north -| dl-network1);
\end{tikzpicture}
}
}
        \hspace{2em}
 	    \subfloat[Multi-task ensemble model\label{fig:approach:ens}]
 	    {%
        \resizebox{0.48\textwidth}{!}{
 			\begin{tikzpicture}
\node[in-out, node distance=1cm] (ens-class) {Task \textit{1}};
\node[in-out, node distance=2.3cm, anchor=south] (ens-intensity) [right=of ens-class] {Task \textit{N}};

\node[process-small, node distance=1cm] (ens-dense1) [below=of ens-class] {Dense Layer};
\node[process-small, node distance=2.3cm, anchor=south] (ens-dense2) [right=of ens-dense1] {Dense Layer};

\path (ens-class.south east) -- (ens-intensity.south west) node[process-huge] (ens-mlp) [midway, below=3.0cm] {\LARGE Multi-Layer Perceptron};

\node[in-out-large, node distance=2.0cm] (ens-lstm) [below=of ens-mlp.south west, anchor=west] {\textit{Task-aware}\\LSTM\\Representation};
\node[in-out-large, node distance=2.0cm] (ens-svr) [below=of ens-mlp.south east, anchor=east] {Feature\\Representation};

\node[in-out-large, node distance=0.3cm] (ens-cnn) [right=of ens-lstm] {\textit{Task-aware}\\CNN\\Representation};
\node[in-out-large, node distance=0.3cm] (ens-gru) [left=of ens-svr] {\textit{Task-aware}\\GRU\\Representation};

\path [draw, line width=0.5mm, black!60!blue, ->] (ens-lstm.north) -- (ens-mlp.south -| ens-lstm);
\path [draw, line width=0.5mm, black!60!blue, ->] ([xshift=0.7cm]ens-lstm.north) -- ([xshift=0.7cm]ens-mlp.south -| ens-lstm);
\path [draw, line width=0.5mm, black!60!blue, ->] ([xshift=-0.7cm]ens-lstm.north) -- ([xshift=-0.7cm]ens-mlp.south -| ens-lstm);

\path [draw, line width=0.5mm, black!60!blue, ->] (ens-cnn.north) -- (ens-mlp.south -| ens-cnn);
\path [draw, line width=0.5mm, black!60!blue, ->] ([xshift=0.7cm]ens-cnn.north) -- ([xshift=0.7cm]ens-mlp.south -| ens-cnn);
\path [draw, line width=0.5mm, black!60!blue, ->] ([xshift=-0.7cm]ens-cnn.north) -- ([xshift=-0.7cm]ens-mlp.south -| ens-cnn);

\path [draw, line width=0.5mm, black!60!blue, ->] (ens-gru.north) -- (ens-mlp.south -| ens-gru);
\path [draw, line width=0.5mm, black!60!blue, ->] ([xshift=0.7cm]ens-gru.north) -- ([xshift=0.7cm]ens-mlp.south -| ens-gru);
\path [draw, line width=0.5mm, black!60!blue, ->] ([xshift=-0.7cm]ens-gru.north) -- ([xshift=-0.7cm]ens-mlp.south -| ens-gru);

\path [draw, line width=0.5mm, black!60!blue, ->] (ens-svr.north) -- (ens-mlp.south -| ens-svr);
\path [draw, line width=0.5mm, black!60!blue, ->] ([xshift=0.7cm]ens-svr.north) -- ([xshift=0.7cm]ens-mlp.south -| ens-svr);
\path [draw, line width=0.5mm, black!60!blue, ->] ([xshift=-0.7cm]ens-svr.north) -- ([xshift=-0.7cm]ens-mlp.south -| ens-svr);

\path [draw, line width=0.5mm, black!60!blue, <-] (ens-class.south) -- (ens-dense1.north -| ens-class);
\path [draw, line width=0.5mm, black!60!blue, <-] (ens-intensity.south) -- (ens-dense2.north -| ens-intensity);

\path [draw, line width=0.5mm, black!60!blue, <-] (ens-dense1.south) -- (ens-mlp.north -| ens-dense1);
\path [draw, line width=0.5mm, black!60!blue, <-] ([xshift=1cm]ens-dense1.south) -- ([xshift=1cm]ens-mlp.north -| ens-dense1);
\path [draw, line width=0.5mm, black!60!blue, <-] ([xshift=-1cm]ens-dense1.south) -- ([xshift=-1cm]ens-mlp.north -| ens-dense1);

\path [draw, line width=0.5mm, black!60!blue, <-] (ens-dense2.south) -- (ens-mlp.north -| ens-dense2);
\path [draw, line width=0.5mm, black!60!blue, <-] ([xshift=1cm]ens-dense2.south) -- ([xshift=1cm]ens-mlp.north -| ens-dense2);
\path [draw, line width=0.5mm, black!60!blue, <-] ([xshift=-1cm]ens-dense2.south) -- ([xshift=-1cm]ens-mlp.north -| ens-dense2);

\end{tikzpicture}
     	}
        }}
        \caption{Proposed Multi-task framework.}
        \label{fig:approach}
\end{figure}
\subsection{Deep Learning Models}\label{dl-models}
We employ the architecture of Figure \ref{fig:approach:ind} to train and tune all the deep learning models using pre-trained GloVe (common crawl 840 billion) word embeddings \cite{2014glove}. In our CNN model, we use two convolution layers followed by two max-pool layers (\textit{conv-pool-conv-pool}). Each convolution layer has 100 filters sliding over 2, 3 and 4 words in parallel. For LSTM/GRU models, we use two stacked LSTM/GRU layers, each having 128 neurons. The CNN, LSTM and GRU layers are followed by two fully connected layers and the output layer. We use 128 (\textit{color coded green} in Figure \ref{fig:approach:ind}) and 100 (\textit{color coded blue} `Dense Layer' in Figure \ref{fig:approach:ind}) neurons in the fully connected layers for all the models. The output layer has multiple neurons depending on the number of tasks in the multi-task framework. The fully connected layer activation is set to \textit{rectified linear} \cite{glorot2011deep}, and the output layer activation is set according to the task - \textit{softmax} for classification \& \textit{sigmoid} for regression. We apply 25\% \textit{Dropout} \cite{hinton2012improving} in the fully-connected layers as a measure of regularization. The \textit{Adam} \cite{adam} optimizer with default parameters is used for gradient based training.

\subsection{Hand-Crafted Feature Vector}\label{feat}
$\bullet$ \textbf{Word and Character Tf-Idf:} Word Tf-Idf weighted counts of 1, 2, 3 grams and character Tf-Idf weighted counts of 3, 4 and 5 grams. \\
$\bullet$ \textbf{TF-Idf Weighted Word Vector Averaging:} Word embeddings models are generally good at capturing semantic information of a word. However, every word is not equally significant for a specific problem. Tf-Idf assigns weights to the words according to their significance in the document. We scale the embeddings of words in the text w.r.t their Tf-Idf weights and use this weighted embedding average of words to create a set of features. \\
$\bullet$ \textbf{Lexicon Features:} \\
- count of positive and negative words using the MPQA subjectivity lexicon \cite{weibe-lexicon} and Bing Liu lexicon \cite{ding2008holistic}. \\
- positive, negative scores from Sentiment140, Hashtag Sentiment lexicon \cite{NRC2}, AFINN \cite{nielsen2011new} and Sentiwordnet \cite{baccianella2010sentiwordnet}. \\
- aggregate scores of hashtags from NRC Hashtag Sentiment lexicon \cite{NRC2}. \\
- count of the number of words matching each emotion from the NRC Word-Emotion Association Lexicon \cite{Mohammad13}. \\
- Sum of emotion associations in NRC-10 Expanded lexicon \cite{bravo2016determining}, Hashtag Emotion Association Lexicon \cite{mohammad2015using} and NRC Word-Emotion Association Lexicon \cite{Mohammad13}. \\
- Positive and negative scores of the emoticons obtained from the AFINN project \cite{nielsen2011new}. \\
$\bullet$ \textbf{Vader Sentiment:} We use Vader sentiment \cite{vader2014} which generates a compound sentiment score for a sentence between -1 (extreme negative) and +1 (extreme positive). It also produces ratio of positive, negative and neutral tokens in the sentence. We use the score and the three ratios as features in our feature based model.

Since the feature vector dimension is too large in comparison with DL representation during ensemble, we project the feature vector to smaller dimension (i.e. 128) through a small MLP network.

\section{Datasets, Experiments and Analysis}
\label{exp}
\subsection{Dataset}
We evaluate our proposed model on the benchmark datasets of WASSA-2017 shared task on emotion intensity (EmoInt-2017) \cite{mohammad:WASSA2017}, EmoBank \cite{emobank:2017} and Facebook posts \cite{fbpost:WASSA2016} for the \textit{coarse-grained emotion analysis}, \textit{fine-grained emotion analysis} and \textit{fine-grained sentiment analysis}, respectively. The dataset of EmoInt-2017 \cite{mohammad:WASSA2017} contains generic tweets representing four emotions i.e. \textit{anger}, \textit{fear}, \textit{joy} and \textit{sadness} and their respective intensity scores. It contains 3613, 347 \& 3142 generic tweets for training, validation and testing, respectively. The EmoBank dataset \cite{emobank:2017} comprises of 10,062 tweets across multiple domains (e.g. \textit{blogs}, \textit{new headlines}, \textit{fiction} etc.). Each tweet has three scores representing \textit{valence}, \textit{arousal} and \textit{dominance} of emotion \textit{w.r.t.} the writer's and reader's perspective. Each score has continuous range of \textit{1} to \textit{5}. For experiments, we adopt 70-10-20 split for training, validation and testing, respectively. The Facebook posts dataset \cite{fbpost:WASSA2016} has 2895 social media posts. Posts are annotated on a nine-point scale with valence and arousal score for sentiment analysis by two psychologically trained annotators. We perform \textit{10-fold} cross-validation for the evaluation. 

\begin{table*}[ht!]
\centering
\subfloat[\textbf{Coarse-grained emotion analysis}: Intensity is on the scale of 0 to 1 \cite{mohammad:WASSA2017}. \label{tab:coarse:emo}]
{
\resizebox{0.465\textwidth}{!}{
\begin{tabular}{|p{16em}|c|c|}
\hline 
\textbf{Text} & \textbf{Emotion} & \textbf{Intensity} \\ \hline \hline
\textit{Just died from laughter after seeing that.} & Joy & 0.92 \\ \hline
\textit{My uncle died from cancer today...RIP.} & Sadness & 0.87 \\ \hline
\textit{Still salty about that fire alarm at 2am this morning.} & Fear & 0.50 \\ \hline
\textit{Happiness is the best revenge.} & Anger & 0.25 \\ \hline
\end{tabular}}
}
\hspace{1em}
\subfloat[\textbf{Fine-grained emotion analysis}: Valence, arousal \& dominance are on the scale of 1 to 5 \cite{emobank:2017}.\label{tab:fine:emo}]
{
\resizebox{0.475\textwidth}{!}{
\begin{tabular}{|p{12em}|c|c|c|}
\hline
\textbf{Text} & \textbf{Valence} & \textbf{Arousal}  & \textbf{Dominance} \\ \hline \hline
\textit{I am thrilled with the price.} & 4.4 & 4.4 & 4.0 \\ \hline
\textit{I hate it, despise it, abhor it!} & 1.0 & 4.4  & 2.2 \\ \hline
\textit{Collision on icy road kills 7.} & 1.2 & 4.2  & 2.2 \\ \hline
\textit{I was feeling calm and private that night.} & 3.2 & 1.6 & 3.0 \\ \hline
\textit{I just hope they keep me here.} & 2.7 & 2.7  & 2.0 \\ \hline
\textit{James Brown's 5-year-old son left out of will.	} & 1.0 & 2.6  & 2.2 \\ \hline
\end{tabular}}
}
\\
\subfloat[\textbf{Fine-grained sentiment analysis}: Valence and arousal are on the scale of 1 to 9 \cite{fbpost:WASSA2016}.\label{tab:fine:sent}]
{
\resizebox{0.985\textwidth}{!}{
\begin{tabular}{|p{40em}|c|c|}
\hline
\textbf{Text} & \textbf{Valence} & \textbf{Arousal} \\ \hline \hline
\textit{I bought my wedding dress Monday and I cant wait to have it on again!!!! its sooo beautiful.} & 8.0 & 8.0 \\ \hline
\textit{Happy, got new friends, and lifes getting smoother.} & 8.0 & 1.5 \\ \hline
\textit{At least 15 dead as Israeli forces attack Gaza aid ships!!!!!!! i hhhhhhate israil} & 1.5 & 8.0 \\ \hline
\textit{The worst way to miss someone is when they r right beside u and yet u know u can never have them.} & 2.5 & 1.5 \\ \hline
\end{tabular}
}
}
\caption{Multi-task examples of emotion analysis and sentiment analysis from benchmark datasets.
\textbf{\textit{Valence}} $\Rightarrow$ Concept of polarity (pleasant / unpleasant); \textbf{\textit{Arousal or Intensity}} $\Rightarrow$ Degree of emotion/sentiments; \textbf{\textit{Dominance}} $\Rightarrow$ Control over a situation;}
\label{exm}
\end{table*}

Few example scenarios for the problems of emotion analysis (coarse-grained \& fine-grained) and sentiment analysis (fine-grained) are depicted in Table \ref{exm}. In the first example shown in Table \ref{tab:coarse:emo}, emotion \textit{`joy'} is derived from the phrase `\textit{died from laughter}' which is intense. However, the emotion associated with the second example which contains similar phrase `\textit{died from cancer}' is \textit{`sadness'}. The third example expresses \textit{`fear'} with mild intensity, whereas, the fourth example conveys \textit{`anger'} emotion with relatively lesser intensity. 

Examples of fine-grained emotion analysis are listed in Table \ref{tab:fine:emo}. Each text is associated with psychologically motivated VAD (\textit{Valence}, \textit{Arousal} \& \textit{Dominance}) scores. \textit{Valence} is defined by pleasantness (positive) or unpleasantness (negative) of the situations. \textit{Arousal} reflects the degree of emotion, whereas, \textit{Dominance} suggests the degree of control over a particular situation. Similarly, Table \ref{tab:fine:sent} depicts the example scenarios for fine-grained sentiment analysis. 

\subsection{Experimental Setup and Results} 
We use Python based libraries, Keras and Scikit-learn for implementation. For evaluation, we compute \textit{accuracy} for the classification (\textit{emotion class}) and \textit{pearson correlation coefficient} for the regression (e.g. \textit{intensity}, \textit{valence}, \textit{arousal} \& \textit{dominance}). Pearson correlation coefficient measures the linear correlation between the actual and predicted scores. The choice of these metrics was inspired from \cite{mohammad:WASSA2017} and \cite{fbpost:WASSA2016}. We normalize the \textit{valence}, \textit{arousal} and \textit{dominance} scores on a \textit{0} to \textit{1} scale. For prediction, we use \textit{softmax} for classification and \textit{sigmoid} for regression. 

\begin{figure*}
\includegraphics[width=1.0\textwidth]{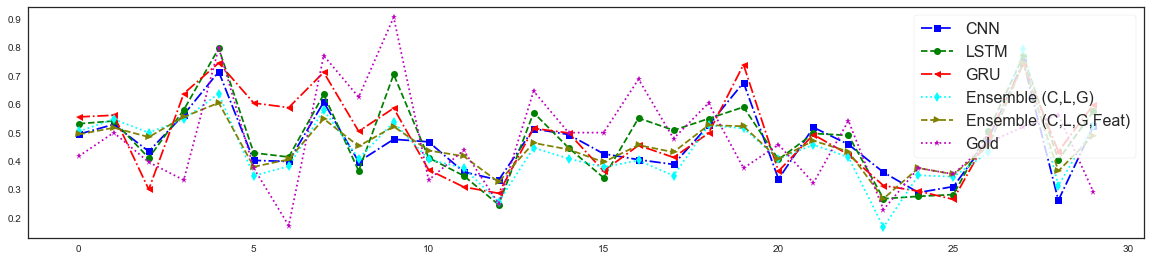}
\caption{Contrasting nature of the individual models and improved scores after ensemble for emotion intensity prediction. \textbf{X-axis}: 30 random samples from the test set. \textbf{Y-axis}: Intensity values.}
\label{emo:graph}
\end{figure*}

Table \ref{result-emo-test} shows the results on the test set for coarse-grained emotion analysis. In multi-task framework, we predict emotion class and intensity together, whereas in single-task framework we build two separate models, one for classification and one for intensity prediction. We follow a dependent evaluation\footnote{Please note that we adopted dependent evaluation strategy as this is commonly used for the evaluation of related-tasks in multi-task framework.} technique where we compute the scores of only those instances which are correctly predicted by the emotion classifier. Such evaluation is informative and realistic as predicting intensity scores for the misclassified instances would not convey the correct information. For direct comparison, we also adopted a similar approach for intensity prediction evaluation in the single-task framework.   
\begin{table*}[h!]
\begin{center}
\resizebox{0.7\textwidth}{!}
{
\begin{tabular}{lcc||cc}
\hline 
\multirow{3}{*}{\bf Models} & \multicolumn{2}{c||}{\bf Multi-task learning} & \multicolumn{2}{|c}{\bf Single-task learning} \\ \cline{2-5}
& {\bf Emotion Class} & {\bf Intensity*} & {\bf Emotion Class} & {\bf Intensity*} \\ \cline{2-5}
& \textbf{Accuracy \%} & \textbf{Pearson} & \textbf{Accuracy \%} & \textbf{Pearson} \\ \hline \hline
CNN (C) & 80.52 & 0.578 & 79.56 & 0.493 \\  
LSTM (L) & 84.69 & 0.625 & 84.02 & 0.572 \\ 
GRU (G) & 84.94 & 0.606 & 83.45 & 0.522 \\ \hline
\hline
Ensemble (C, L \& G) & 85.93 & 0.657 & 85.77 & 0.596 \\ 
Ensemble (C, L, G \& Feat) & \textbf{89.88} & \textbf{0.670} & 89.52 & 0.603 \\ \hline
\textit{Significance T-test (p-values)}\footref{significance} & \textit{0.073} & \textit{0.001} & - & - \\ \hline
\end{tabular}
}
\end{center}
\caption{\textbf{Coarse-grained Emotion Analysis:} Experimental results for multi-task (\textit{i.e. single model for both tasks in parallel}) and single-task (\textit{i.e. first a tweet is classified to an emotion class and then intensity is predicted }) learning framework for \textbf{EmoInt-2017 datasets} \cite{mohammad:WASSA2017}. Significance \textit{T}-test (p-values) are \textit{w.r.t.} single task learning.
}
\label{result-emo-test}
\end{table*}
The first half of Table \ref{result-emo-test}
reports the evaluation results for three deep learning models. In multi-task framework, CNN reports 80.52\% accuracy for classification and 0.578 Pearson score for intensity prediction. The multi-task LSTM and GRU models obtain 84.69\% \& 84.94\% accuracy values and 0.625 \& 0.606 Pearson scores, respectively. The corresponding models in single-task framework report 79.56\%, 84.02\% \& 83.45\% accuracy values and 0.493, 0.572 \& 0.522 Pearson scores for CNN, LSTM \& GRU models, respectively. 
It is evident that multi-task models perform better than the single-task models by a convincingly good margin for intensity prediction, and better for class prediction. On further analysis, we observe that these models obtain quite similar performance numerically. However, they are quite contrasting on a qualitative side. Figure \ref{emo:graph} shows the contrasting nature of different individual models for emotion intensity. In some cases, prediction of one model is closer to the gold intensity than the other models and vice-versa. We observe similar trends for the other tasks as well. An ensemble system constructed using only deep learning models achieves the enhanced accuracy of 85.83\% and Pearson score of 0.657. Further inclusion of hand-crafted feature vectors (c.f. section \ref{feat}) in the ensemble network results in an improvement of around 4\% accuracy and 1.5\% Pearson score.   

We report the results for \textit{fine-grained emotion} \& \textit{sentiment analysis} in Table \ref{result-emobank-fb-test}. Similar to \textit{coarse-grained emotion analysis} we observe that multi-task models achieve the improved Pearson scores (0.635, 0.375 \& 0.277) as compared to the single-task based models (0.616, 0.355 \& 0.234) for the three tasks, i.e. valence, arousal and dominance, respectively. The ensemble approach also achieves better performance compared to each of the base models for all the tasks. For fine-grained sentiment analysis, deep learning based models i.e. CNN, LSTM \& GRU obtain Pearson scores of 0.678, 0.671 \& 0.668 for \textit{valence} in multi-task environment. The ensemble of these three models and hand-crafted feature representation via MLP obtains an increased Pearson score of 0.727. The proposed approach also achieves the best Pearson score of 0.355 for \textit{arousal}.   
\begin{table*}[ht!]
\begin{center}
\resizebox{0.9\textwidth}{!}
{
\begin{tabular}{lccc|ccc||cc|cc}
\hline 
\multirow{3}{*}{\bf Models} & \multicolumn{6}{c||}{\bf Emotion Analysis - EmoBank} & \multicolumn{4}{|c}{\bf Sentiment Analysis - FB post} \\ \cline{2-11}
& \multicolumn{3}{c|}{\bf Multi-task} & \multicolumn{3}{c||}{\bf Single-task} & \multicolumn{2}{c|}{\bf Multi-task} & \multicolumn{2}{c}{\bf Single-task} \\ \cline{2-11}
& {\bf Val} & {\bf Aro} & {\bf Dom} & {\bf Val} & {\bf Aro} & {\bf Dom} & {\bf Val} & {\bf Aro} & {\bf Val} & {\bf Aro} \\ \hline \hline
CNN (C) & 0.567 & 0.347 & 0.234 & 0.552 & 0.334 & 0.222 & 0.678 & 0.290 & 0.666 & 0.283 \\
LSTM (L) & 0.601 & 0.337 & 0.245 & 0.572 & 0.318 & 0.227 & 0.671 & 0.324 & 0.655 & 0.315\\
GRU (G) & 0.569 & 0.315 & 0.243 & 0.553 & 0.306 & 0.227 & 0.668 & 0.313 & 0.657 & 0.294 \\ \hline \hline
Ensemble (C, L \& G) & 0.618 & 0.365 & 0.263 & 0.603 & 0.351 & 0.234 & 0.695 & 0.336 & 0.684 & 0.324 \\ 
Ensemble (C, L, G \& Feat) & \textbf{0.635} & \textbf{0.375} & \textbf{0.277} & 0.616 & 0.355 & 0.237 & \textbf{0.727} & \textbf{0.355} & 0.713 & 0.339 \\ \hline
\textit{Significance T-test (p-values)}\footnotemark & \textit{0.048} & \textit{0.027} & \textit{0.310} & - & - & - & \textit{0.033} & \textit{0.024} & - & - \\ \hline
\end{tabular}}
\end{center}
\caption{\textbf{Fine-grained Emotion and Sentiment Analysis:} Experimental results for multi-task and single-task learning framework on \textbf{EmoBank datasets} \cite{emobank:2017} \& \textbf{FB post datasets} \citep{fbpost:WASSA2016}. \textbf{\textit{Val}:} Valence, \textbf{\textit{Aro}:} Arousal \& \textbf{\textit{Dom}:} Dominance. Significance \textit{T}-test (p-values) are \textit{w.r.t.} single task learning. 
}
\label{result-emobank-fb-test}
\end{table*}
\footnotetext{\label{significance}We generate 20 random samples on normal distribution curve with $\sigma=0.05$}


We observe two phenomenon from these results: \textbf{a)} use of multi-task framework for related tasks indeed helps in achieving generalization; and \textbf{b)} the ensemble network leverages the learned representations of three base models \& the feature vector and produces superior results.

\begin{table}[ht!]
\centering
\resizebox{0.48\textwidth}{!}
{
\begin{tabular}{p{10em} c C{4.5em} C{5em}}
\hline 
\multirow{2}{*}{\bf Models} & \multirow{2}{*}{\bf Emotion} & \multicolumn{2}{c}{\bf Emotion Intensity} \\ \cline{3-4}
& \textbf{Class} & Dependent Evaluation & Independent Evaluation \\ \hline \hline 

Baseline$^+$ & - & - & 0.648 \\
Prayas (Multi-task)* & - & - & 0.662\\
\hline
Proposed (Single-task) & 89.52 & - & 0.603 \\ 
Proposed (Multi-task) & \textbf{89.88} & \textbf{0.670} & 0.647 \\
\hline 
\end{tabular}
}
\caption{\textbf{Coarse-grained Emotion Analysis:} Comparative results. \textbf{*}Prayas \cite{jain-EtAl:2017:WASSA2017} was the top system at EmoInt-2017. They treated intensity prediction of four emotion classes as multi tasks; \textbf{Dependent evaluation:} Intensity was evaluated following emotion classification; \textbf{Independent evaluation:} Intensity score is evaluated independent of the emotion class; $\mathbf{^+}$Baseline system is taken from \cite{mohammad:WASSA2017}. \textit{Please note that both the Prayas and baseline systems have different setups than the proposed method and do not provide an ideal scenario for direct comparison.}
}
\label{result-compare-emo-coarse}
\end{table}

\begin{table}[ht]
\centering
\resizebox{0.48\textwidth}{!}
{
\begin{tabular}{l c c }
\hline 
\multirow{2}{*}{\bf Models} & \multirow{1}{*}{\bf Valence} & \multicolumn{1}{c}{\bf Arousal}\\ \cline{2-3}
& \multicolumn{2}{c}{\textbf{Pearson}} \\
& \multicolumn{2}{c}{\textbf{correlation}} \\ \hline \hline
System \cite{fbpost:WASSA2016} & 0.650 & \textbf{0.850} \\
System - X* & 0.390 & 0.105 \\ \hline
Proposed (Single-task) & 0.713 & 0.339 \\ 
Proposed (Multi-task) & \textbf{0.727} & 0.355 \\ \hline 
\end{tabular}
}
\caption{\textbf{Fine-grained Sentiment Analysis:} Comparative results for Facebook posts dataset. \textbf{System - X*}: \textit{Google search lists this paper in the citation list of \cite{fbpost:WASSA2016}, however, the publication details are not available. The pdf is available at \url{www.goo.gl/DcdaHF}.
}} 
\label{result-compare-sent-fine}
\end{table}

\subsection{Comparative Analysis}
For \textit{coarse-grained sentiment analysis}, we compare our proposed approach with Prayas system \cite{jain-EtAl:2017:WASSA2017}, which was the top performing system at EmoInt-2017 \cite{mohammad:WASSA2017} shared task on Emotion Intensity. Prayas \cite{jain-EtAl:2017:WASSA2017} used an ensemble of five different neural network models including a multitasking feed-forward model. Although the final model was built for each emotion type separately, in multi-task model the authors treated four emotion classes as the four tasks. However, our proposed approach treats emotion classification and emotion intensity prediction as two separate tasks, and then learns jointly (a completely different setup than Prayas).
Prayas reported the Pearson score of 0.662 for emotion intensity. In comparison, our proposed approach obtains a Pearson score of 0.670 for dependent evaluation, and 0.647 for independent evaluation. Statistical \textit{T-test} shows that the value (0.679) is statistically significant over the model of Prayas.

Similarly, we do not compare our proposed approach with other systems of EmoInt-2017 \cite{mohammad:WASSA2017} because of the following two reasons: a) those systems are of single-task nature as compared to our proposed multi-task; and b) separate models were trained for each of the emotions and an average score was reported as compared to a unified single model that addressed all the emotions and their intensity values altogether. The baseline system for emotion intensity prediction in Table \ref{result-compare-emo-coarse} is taken from \newcite{mohammad:WASSA2017}, which also differs from our proposed approach \textit{w.r.t.} the above two points, and hence does not provide an ideal candidate for direct comparison.
\begin{table*}[ht!]
\centering
\begin{minipage}[l]{0.20\textwidth}
\resizebox{1.0\textwidth}{!}
{
\begin{tabular}{|c|c|c|c|}
\hline
15 & 9 & 40 & \cellcolor{blue!25} 548 \\ \hline
16 & 17 & \cellcolor{blue!25} 901 & 80 \\ \hline
11 & \cellcolor{blue!25} 657 & 25 & 12 \\ \hline
\cellcolor{blue!25} 718 & 31 & 29 & 33 \\ \hline
\multicolumn{1}{c}{\small Anger} & \multicolumn{1}{c}{\small Joy \qquad}  & \multicolumn{1}{c}{\small Fear} & \multicolumn{1}{c}{\small Sadness} \\
\end{tabular}}
\caption{Confusion matrix for EmoInt-2017 emotion classification problem.}
\label{tab:confusion}
\end{minipage}
\hspace{1em}
\begin{minipage}[l]{0.75\textwidth}
\resizebox{1.0\textwidth}{!}
{
\begin{tabular}{c p{26em} c c l} \hline
& \textbf{Text} & \textbf{Actual} & \textbf{Predicted} & \textbf{Possible Reason} \\ \hline

\multirow{4}{*}{\rot{EmoInt-2017}} & \multicolumn{4}{l}{\textbf{Emotion Classification}} \\ 
& \textit{Going back to \textbf{blissful ignorance}.} & Sad & Joy & Metaphoric sentence. \\
\cline{2-5}
& \multicolumn{4}{l}{\textbf{Intensity Prediction}} \\
& \textit{Never let the sadness of your past \textbf{ruin} your future.} & Sad/0.29 & Sad/0.64 & Strong expression. \\ \hline \hline
\multirow{6}{*}{\rot{EmoBank}} & \multicolumn{4}{l}{\textbf{Valence Prediction}} \\ 
& \textit{It's \textbf{summertime}, so it must be time for \textbf{CAMP!}} & 4.4 & 3.1 & Implicit emotion. \\ \cline{2-5}

& \multicolumn{4}{l}{\textbf{Arousal Prediction}} \\
& \textit{The company is \textbf{on a roll.}} & 4.0 & 2.8 & Implicit emotions. \\ \cline{2-5}
& \multicolumn{4}{l}{\textbf{Dominance Prediction}} \\ 
& \textit{Three days later, another \textbf{B-29} from the \textbf{509th} bombed Nagasaki.} & 2.0 & 3.3 & \multirow{1}{*} {Numerical entities.} \\  
\hline \hline

\multirow{4}{*}{\rot{FB Posts}} & \multicolumn{4}{l}{\textbf{Valence Prediction}} \\ 
& \textit{I am on \textbf{cloud nine} right now.} & 7.5 & 4.3 & \multirow{1}{*}{Idiomatic expressions.} \\ 
\cline{2-5}
& \multicolumn{4}{l}{\textbf{Arousal Prediction}} \\
& \textit{\textbf{Thank you} all for \textbf{wishing} me a \textbf{happy birthday.}} & 1.5 & 8.1 & \multirow{1}{*}{Strong expressions.} \\  
\hline
\end{tabular}
}
\caption{\textbf{Error Analysis:} Frequent error cases for the best performing multi-task models.}
\label{tab-error-extensive}
\end{minipage}
\end{table*}

We do not compare emotion classification tasks with other systems as we could not find any related works on the same dataset. A comparative analysis is presented in Table \ref{result-compare-emo-coarse}. It is evident that solving both the tasks together in a multi-task setting produces better performance than solving these two tasks separately in a single-task setting.

The datasets for \textit{fine-grained emotion analysis} and \textit{fine-grained sentiment analysis} problems i.e. EmoBank \cite{emobank:2017} and Facebook posts \cite{fbpost:WASSA2016} are relatively recent datasets and limited studies are available on these. We did not find any existing system that evaluated Pearson score for these datasets except the resource paper of Facebook posts \cite{fbpost:WASSA2016}. For \textit{valence} in \textit{fine-grained sentiment analysis} a Pearson score of 0.650 has been reported in \cite{fbpost:WASSA2016} using a Bag-of-Words (BoW) model. In comparison, our proposed approach reports the Pearson score of 0.727, an improvement of 7 points. For \textit{arousal} Preo\c{t}iuc-Pietro et al. \shortcite{fbpost:WASSA2016} reported a Pearson score of 0.850 as compared to 0.355 of ours. It should be noted that we tried to reproduce the scores of \cite{fbpost:WASSA2016} using the same BoW model. We obtained the similar Pearson score of 0.645 for \textit{valence}, however, we could not reproduce the reported results for arousal (we obtained Pearson score of 0.27)\footnote{Please note that the research reported in \url{www.goo.gl/DcdaHF} obtained only 0.105 on the same dataset}. In Table \ref{result-compare-sent-fine}, we demonstrate the comparative results for \textit{fine-grained sentiment analysis}. 

\subsection{Error Analysis}
We perform qualitative error analysis on the predictions of our best performing multi-task models. At first, we identify the most commonly occurring errors and then we analyze 15 test instances for each such error to detect the common error patterns. A number of frequently occurring error cases along with their possible reasons are shown in Table \ref{tab-error-extensive}. We observe that the main sources of errors are metaphoric sentences, strong expressions, implicit emotions and idiomatic expressions.

We also compare the predictions of multi-task models against single-task models. We observe that in many case multi-task learning performs better (correct or closer \textit{w.r.t.} gold labels) than single-task learning. For the emotion classification problem we analyze the confusion matrix and observe that the proposed model often confuses between \textit{fear} and \textit{sadness} class. In total 80 tweets ($\sim$8\%) representing \textit{fear} are misclassified as \textit{sadness}, whereas, 40 instances ($\sim$6.5\%) of \textit{sadness} are misclassified as \textit{fear}. The confusion matrix is depicted in Figure \ref{tab:confusion}. We also perform statistical significance test (\textit{T-test}) on the 10 runs of the proposed approach and observe that the obtained results are significant with $\textnormal{\textit{p}-values}<0.05$ .

\section{Conclusion}
\label{con}
In this work, we have proposed a multi-task ensemble framework for emotion analysis, sentiment analysis and intensity prediction. For ensemble we employed a MLP network that jointly learns multiple related tasks. First, we have developed three individual deep learning models (i.e. CNN, LSTM and GRU) to extract the learned representations. The multi-task ensemble network was further assisted through a hand-crafted feature vector. We evaluate our proposed approach on three benchmark datasets related to sentiment, emotion and intensity. Experimental results show that the multi-task framework is comparatively better than the single-task framework. Emotion detection can also be projected as multi-labeling task. However, due to absence of multi-emotion dataset we do not evaluate the proposed method on multi-emotion task. It should be noted that our model can easily be adapted to multi-label emotion classification. 

\bibliography{reference}
\bibliographystyle{acl_natbib_nourl}
\end{document}